\documentclass[journal,article,submit,pdftex,moreauthors]{Definitions/mdpi} 
\firstpage{1} 
\pubvolume{1}
\issuenum{1}
\articlenumber{0}
\pubyear{2022}
\copyrightyear{2022}
\datereceived{} 
\dateaccepted{} 
\datepublished{} 
\hreflink{https://doi.org/} 

\usepackage{microtype}
\usepackage{slashbox}

\usepackage{times}
\usepackage{latexsym}
\usepackage[T1]{fontenc}
\usepackage[utf8]{inputenc}

\usepackage{amsfonts}
\usepackage{amsmath}
\usepackage{amssymb}        
\usepackage{bold-extra}     
\usepackage{bm}             

\usepackage{graphicx}       
\usepackage{multirow}       
\usepackage{booktabs}       
\usepackage{subcaption}     
\usepackage[hang,flushmargin]{footmisc}  
\usepackage{soul}
\usepackage{ulem}
\usepackage{lipsum}

\usepackage{enumitem}
\usepackage{xcolor}
\sethlcolor{yellow}

\Title{InterviewBot: Real-Time End-to-End Dialogue System to Interview Students for College Admission}

\TitleCitation{InterviewBot: Real-Time End-to-End Dialogue System to Interview Students for College Admission}

\Author{Zihao Wang$^{1,\ast}$, Nathan Keyes$^{2,\bullet}$, Terry Crawford$^{2,\bullet}$ and Jinho D. Choi$^{1,\ast}$\orcidA{}}

\AuthorNames{Zihao Wang, Nathan Keyes, Terry Crawford and Jinho D. Choi}

\AuthorCitation{Wang, Z.; Keyes, N.; Crawford, T.; Choi, J. D.}

\address{%
$^{1}$ \quad Department of Computer Science, Emory University, Atlanta, GA, USA\\
$^{2}$ \quad InitialView, Atlanta GA, USA\\
${\bullet}$ \quad nathan@initialview.com (N.K.), terry@initialview.com (T.C.)
}

\corres{Correspondence: zihao.wang2@emory.edu (Z.W.), jinho.choi@emory.edu (J.C.)}

\abstract{
We present the InterviewBot that dynamically integrates conversation history and customized topics into a coherent embedding space to conduct 10 mins hybrid-domain (open and closed) conversations with foreign students applying to U.S. colleges for assessing their academic and cultural readiness.
To build a neural-based end-to-end dialogue model, 7,361 audio recordings of human-to-human interviews are automatically transcribed, where 440 are manually corrected for finetuning and evaluation.
To overcome the input/output size limit of a transformer-based encoder-decoder model, two new methods are proposed, context attention and topic storing, allowing the model to make relevant and consistent interactions.
Our final model is tested both statistically by comparing its responses to the interview data and dynamically by inviting professional interviewers and various students to interact with it in real-time, finding it highly satisfactory in fluency and context awareness.
}


\begin{document}

\section{Introduction}
\label{sec:introduction}

\noindent With the latest advancement of Conversational AI, end-to-end dialogue systems have been extensively studied \cite{zhang-etal-2020-dialogpt, adiwardana2020towards, roller-etal-2021-recipes}. 
One critical requirement is context awareness; robust dialogue systems must consider relevant parts in conversation history to generate pertinent responses \cite{serban2017hierarchical, mehri2019pretraining, bao-etal-2020-plato, zhou2021eva, xu-etal-2022-beyond}.
However, these systems still suffer from issues such as hallucination, inconsistency, or lacking commonsense \cite{bao2021plato}, hindering them from taking place in real applications.

Numerous admission interviews are given every year to students located in 100+ countries applying to colleges in the U.S., where the interviews are often conducted online. 
Those interviews are usually unscripted, with an emphasis on asking the applicants thought-provoking questions based on their interests and experiences. 
The main objective is to provide decision-makers (e.g., admissions officers, faculty members) with an unfiltered look at those students in a daily academic environment.

Building an interview chatbot, called InterviewBot, will save time and effort for the interviewers and provide foreign students with a cost-efficient way of practicing interviews when native speakers are unavailable.
Nonetheless, there are a few hurdles to developing an end-to-end InterviewBot.
First, it is hard to collect a sufficient amount of data covering dialogues crossing open \& closed domains (Section~\ref{sec:dataset}).
Second, most transformer-based encoder-decoder models adapted by current state-of-the-art systems are not designed to handle long contexts; thus, they often repeat or forget previously discussed topics (Section~\ref{sec:generation-model}).
Third, it is demanding to find appropriate people to interactively test such a dialogue system with the professional objective (Section~\ref{sec:evaluation}).

This paper presents an end-to-end dialogue system that interacts with international applicants to U.S. colleges.
The system questions critical perspectives, follows up on the interviewee's responses for in-depth discussions, and makes natural transitions from one topic to another until the interview ends, which lasts about 30 turns (5 mins for text-based, 10 mins for spoken dialogues).
To the best of our knowledge, it is the first real-time system using a neural model, completely unscripted, conducting such long conversations for admission interviews.
Our technical contributions are summarized as follows:

\begin{itemize}
    \item We have developed a contextualized neural model designed to perform diarization tasks on text transcripts alone.
    \item We have integrated a sliding window technique to overcome the input token limit and restore the completeness of the input in the latent space.
    \item We have integrated extracted topics from the conversation to address issues related to topic repetition, off-topic discussions, and premature endings in conversations.
\end{itemize}

\noindent The remaining sections are organized as follows: Section~\ref{sec:related-work} reviews current dialogue models, their applications, and limitations. Section~\ref{sec:materials_methods} describes datasets and our speaker diarization and InterviewBot model architectures in detail. Section~\ref{sec:evaluation} gives experiment results on diarization and InterviewBot dialogue generation. Section~\ref{sec:appendix} and Section~\ref{sec:conclusion} conduct discussions on the results and conclude the paper.

\section{Related Work}
\label{sec:related-work}



\noindent Dialogue systems can be categorized into closed- and open-domain systems \cite{ilievski2018goal}.
Closed-domain systems require efficient access to domain knowledge \cite{lian2019learning} and serve specific professions such as education \cite{cunningham2019review}, healthcare \cite{fan2021utilization,amiri2022chatbot}, or customer service \cite{baier2018conversational,nichifor2021artificial}.
Open-domain systems converse across multiple domains with natural transitions \cite{adiwardana2020towards} and conduct interactions in a broader horizon \cite{ahmadvand2018emory, wang2017emersonbot, finch2020emora}.
For admission interviews, however, the conversation is often a mixture of closed (job-related questions) and open-domain (general aspects of the applicant) dialogues, which makes it more challenging to build an end-to-end system.

Several dialogue systems have been developed to communicate with humans for information exchange or elicitation across multiple domains \cite{safi2020technical,khoa2021impact,okonkwo2021chatbots}. 
\cite{finch2020emora} built a conversational system to converse proactively on popular topics with Alexa users by providing them with the requested information as well as pre-crafted transitions.
\cite{li2017confiding} established a virtual interviewer to study on the effect of personality on confiding and listening to virtual agents.
\cite{Kim2019Comparing} studies the role of a chatbot in a survey setup. Although these dialogue systems have shown their effectiveness in achieving their goals, they all heavily rely on design templates.
Conversational agents for interviews have been experimented with for law enforcement \cite{minhas2022protecting}, healthcare \cite{ni2017mandy}, job application \cite{xiao2019should}, and psychology \cite{siddig2019psychologist}, among which most are proof of concept. 
A few interview bots have been developed on commercial platforms such as Google Dialogflow and IBM Watson Assistant, with the limitation of pre-scripted interviews; thus, they cannot proactively follow up on the user content.

Context and memory have been studied as key factors to affect model performance in context-heavy settings.
\cite{9681776} proposed a memory transformer to hierarchically employ memory to improve translation performance.
However, in a more complex conversation setup, dialogue flow is not only about correlations between sentences or words in the semantics but rather the proceeding of conversations with a depth of topics and transitions to other topics.
Other models such as \cite{raheja-tetreault-2019-dialogue} and \cite{9533452} have proposed context and external knowledge-based models in conversation-related tasks. 
Although, the effort was proven to improve based on specific metrics, still not sufficient to improve the overall dialogue flow of conversations.

Deep language models, such as Blenderbot \cite{roller-etal-2021-recipes} and Bart \cite{lewis2019bart}, have taken context into consideration. 
However, the limitation on the length of input tokens as well as conversation history has bottlenecked their applications in the real world.
Recent surges of large language models, such as ChatGPT \cite{openai2023gpt4} and LLaMa \cite{touvron2023llama}, have shown strong evidence of improvement with respect to context integration. 
Nevertheless, there are always limitations on the input length, as well as effective ways of integrating different contexts into a language model.



%
\section{Materials and Methods}
\label{sec:materials_methods}

\subsection{Interview Dataset}
\label{sec:dataset}

\noindent Audio recordings of 7,361 interviews are automatically transcribed with speaker identification by the online tool RevAI,\footnote{\url{https://www.rev.ai}} where 440 are manually corrected on speaker ID assignment for finetuning and evaluation of our models (Table~\ref{tab:annotated-data}).
Each recording contains an average of a $\approx$15-min long dialogue between an interviewer and an interviewee.
The interviews were conducted by 67 professionals in 2018 - 2022.
The largest age group of interviewees is 18-years-old with 59.3\%, followed by 17-years-old with 29.4\%.
The male-to-female ratio is 1.2:1.
The major country of origin is China with 81.4\% followed by Belgium with 10.5\%, alongside 37 other countries.
Table \ref{tab:annotated-data} provides detailed demographics of the interviewees.

All recordings are transcribed into text and speakers are identified automatically.
For speech recognition, three tools from Amazon,\footnote{\url{https://aws.amazon.com/transcribe}} Google,\footnote{\url{https://cloud.google.com/speech-to-text}} and RevAI\footnote{\url{https://www.rev.ai}} are assessed on 5 recordings for speaker diarization, achieving the F1-scores of 66.3\%, 50.1\%, and 72.7\%, respectively.\footnote{The same metric as in Table~\ref{tab:diarization-results} is used for this evaluation.}

\begin{table}[htbp!]
\caption{Distributions of our data. D: num of dialogues, U: avg-num of utterances per dialogue, S1/S2: avg-num of tokens per utterance by interviewer/interviewee. \texttt{TRN}/\texttt{DEV}/\texttt{TST}: training/development/evaluation (annotated) sets. \texttt{RAW}: unannotated set (auto-transcribed).}
\vspace{0.5em}
\centering\small { 
\begin{tabular}{crrrr}
\toprule
 & \multicolumn{1}{c}{\bf D} & \multicolumn{1}{c}{\bf U} & \multicolumn{1}{c}{\bf S1} & \multicolumn{1}{c}{\bf S2} \\
\midrule
\tt TRN &   140 & 43.8 & 39.3 & 64.0 \\
\tt DEV & 150 & 45.0 & 36.2 & 60.3 \\
\tt TST &   150 & 44.3 & 37.8 & 61.3 \\
\midrule
\tt RAW & 6,921 & 40.4 & 41.5 & 67.6 \\
\bottomrule
\end{tabular}}
\label{tab:annotated-data}
\end{table}

\begin{figure}[htbp!]
\centering
\includegraphics[width=0.8\columnwidth]{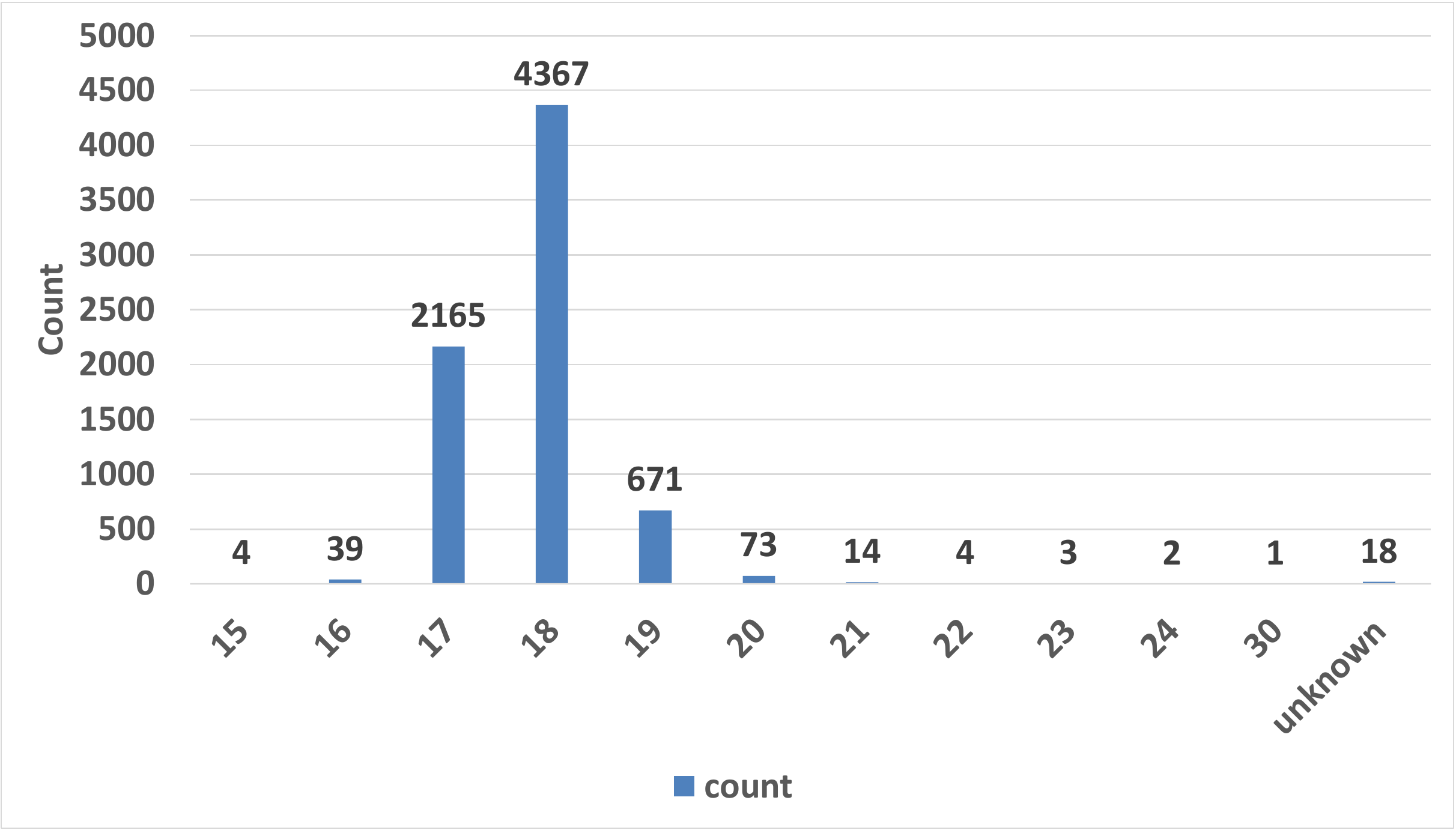}
\caption{The interviewee's age demographics.}
\label{fig:interviewee-age-demographics}
\end{figure}
\begin{figure}[htbp!]
\centering
\includegraphics[width=0.8\columnwidth]{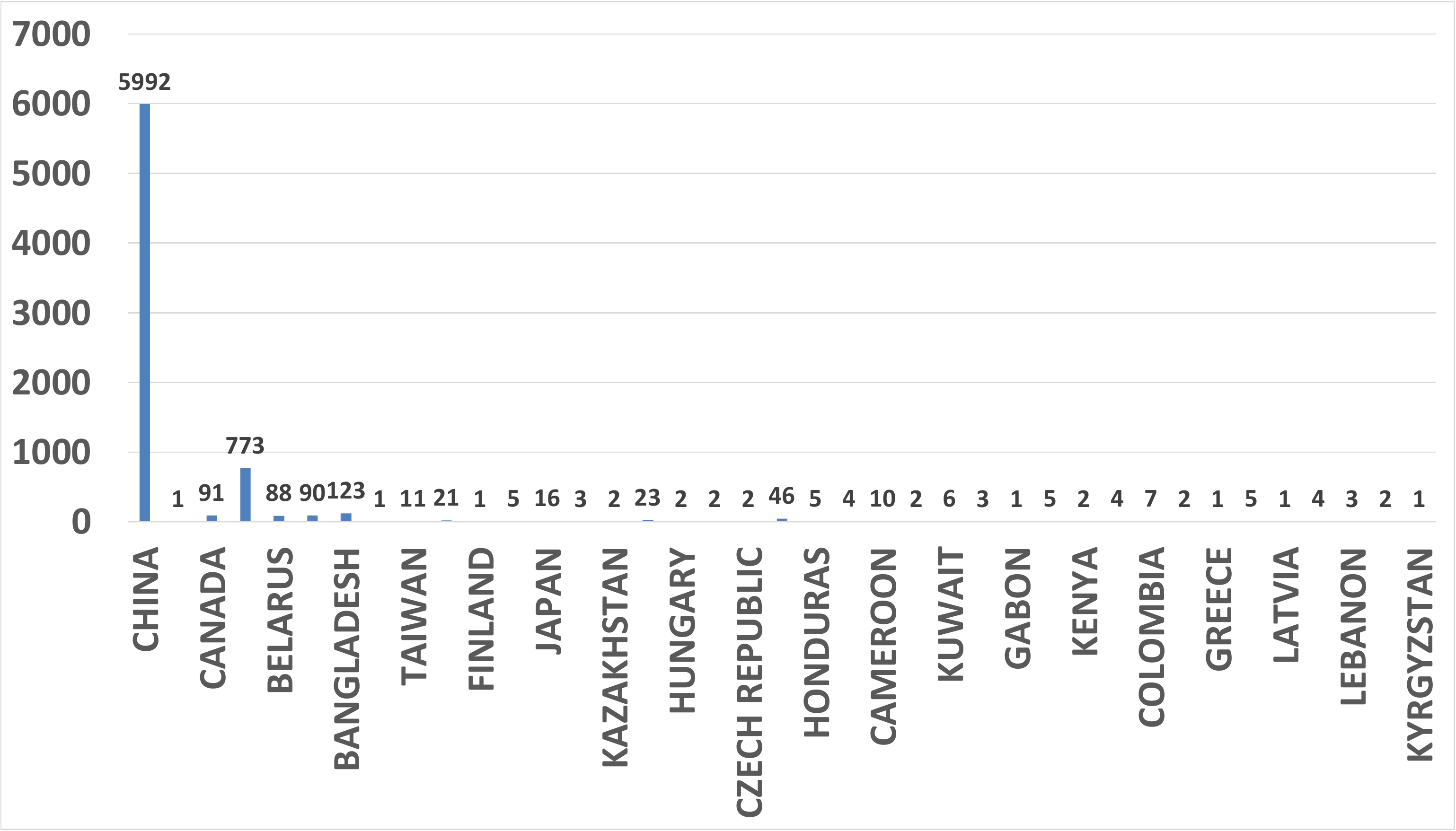}
\caption{The interviewee's country demographics.}
\label{fig:interviewee-country-demographics}
\end{figure}
\begin{figure}[htbp!]
\centering
\includegraphics[width=0.5\columnwidth]{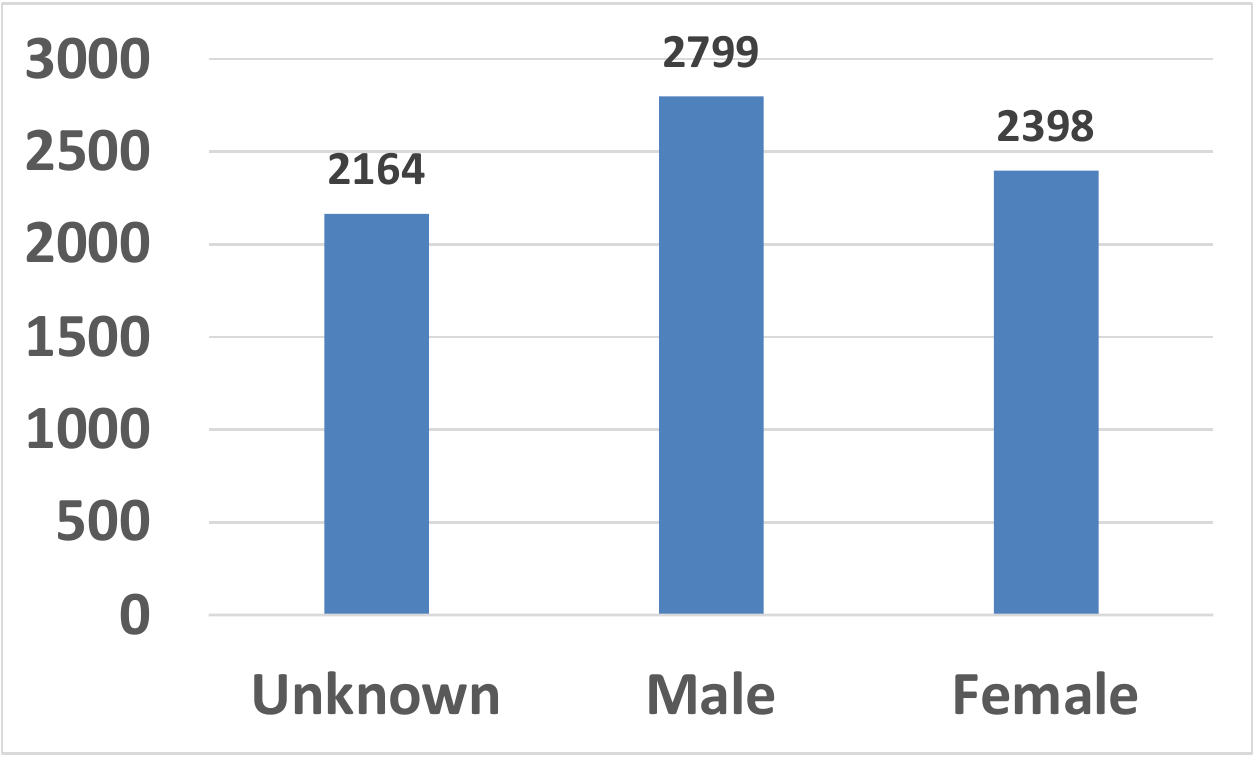}
\caption{The interviewee's gender demographics.}
\label{fig:interviewee-gender-demographics}
\end{figure}

\noindent Figure~\ref{fig:interviewee-age-demographics} shows the distribution of the ages of applicants. 
Most interviewees are between 17~to~19, which is an accurate reflection of the ages of high school students applying to colleges.
Figure~\ref{fig:interviewee-country-demographics} shows the distribution of the applicants' countries of origin.
There are 38 countries in total.
The majority of applicants come from China.
Other major countries are Belgium, Bangladesh, Canada, India, and Belarus.
The gender distribution of applicants is shown in Figure \ref{fig:interviewee-gender-demographics}. 
The numbers of male and female applicants are close, with the exclusion of applicants not providing gender information.

\subsection{Speaker Diarization}
\label{sec:speaker-diarization}

\noindent Speaker diarization is the task of segmenting an audio stream into utterances according to the speaker's identity and is considered critical in automatic transcription \cite{anguera2012speaker}.
Conversation data with diarization errors can lead to a major failure in building robust dialogue models.
Our most accurate transcriber, RevAI, still gives 27.3\% errors for speaker diarization (Section~\ref{sec:dataset}).
The main reason is that audios from the interviewer (\texttt{S1}) and the interviewee (\texttt{S2}) are recorded in one channel, so that they are saved in a single waveform, while no clear pauses exist between \texttt{S1} \& \texttt{S2}'s speeches or their speeches often overlap.
The following example illustrates when the speech of \texttt{S2} (\uwave{underlined}) is not recognized as a separate utterance:

\vspace{1.5ex}
\noindent \texttt{S1}: Hi , it 's nice to meet you . \uwave{Nice to meet you .}\\
\texttt{S2}: Um , can you tell me what is a topic that um , you cannot stop talking about ?
\vspace{1.5ex}

\begin{figure}[htbp!]
\centering
\includegraphics[width=0.7\textwidth]{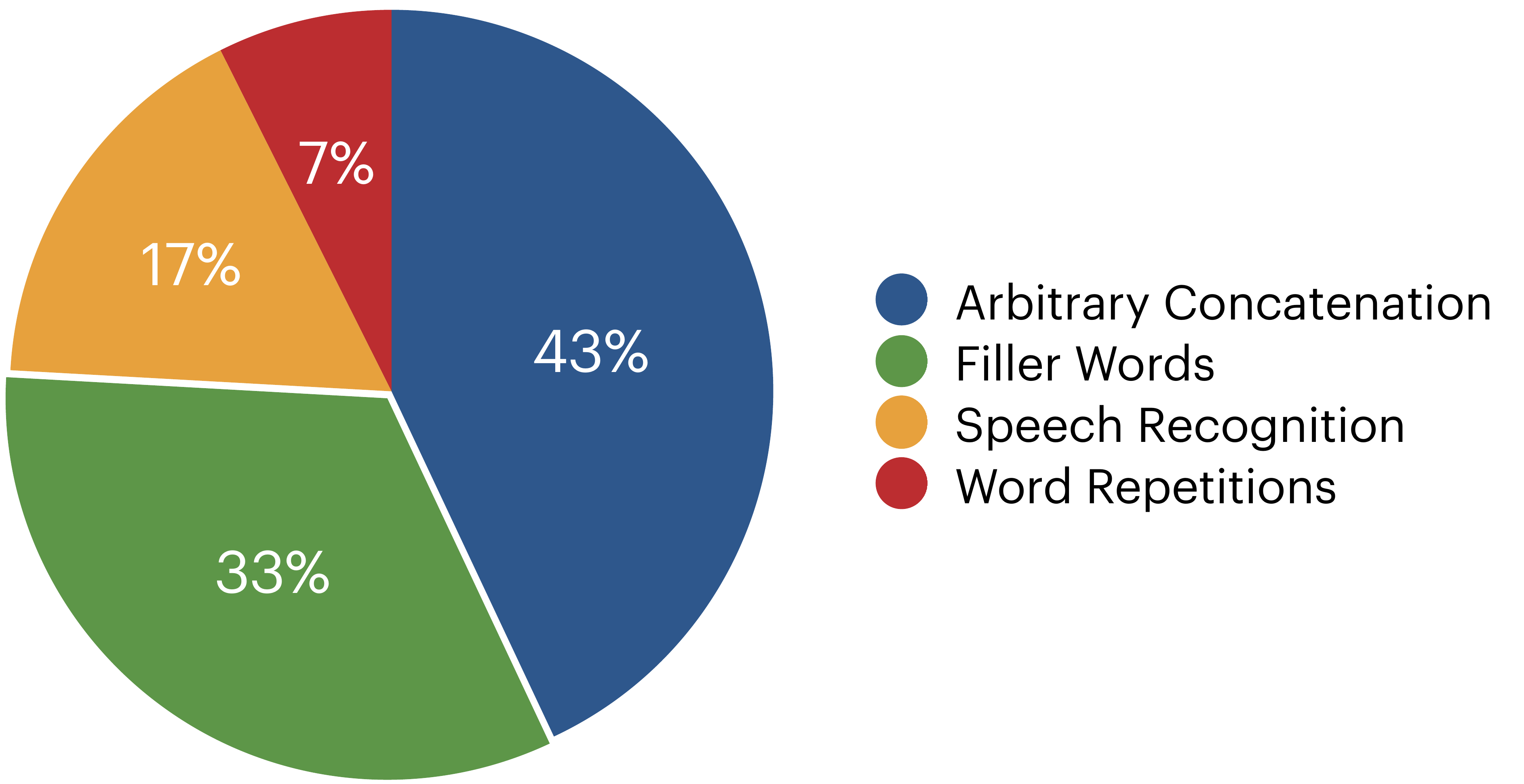}
\caption{Distributions of the diarization error types. Appendix~\ref{app:diarization-error-example} provides examples of each error type.}
\label{fig:diarization-errors}
\end{figure}

\noindent Thus, speaker diarization models are developed to provide clean data to our dialogue model (Sec.~\ref{sec:generation-model}).
Figure~\ref{fig:diarization-errors} depicts the distributions of different types of diarization errors found in 100 dialogues. 
Most errors are caused by filler words and arbitrary concatenation (joining multiple utterances as one with no apparent patterns, not caused by filler words).

\subsubsection{Manual Annotation}
\label{ssec:manual-annotation}

\noindent 440 dialogues are sampled, in which every token is annotated either \texttt{1} if it is one of the last two tokens of an utterance before the speaker is switched, and \texttt{0} otherwise. 
For the above example, the 8-9'th tokens are the last two tokens of the utterance before it switches to \texttt{S2} and so are the 13-14'th tokens before switching to \texttt{S1}; thus, they are annotated~\texttt{1}:\footnote{We also annotated only the last token as \texttt{1}, or annotated all words from \texttt{S0} as \texttt{0} and from \texttt{S1} as \texttt{1}, which yielded worse results in terms of the end performance.}

\vspace{1.5ex}
\begin{tabular}{cccccccccccccc}
Hi & , & it & 's & nice & to & meet & you & . & Nice & to & meet & you & .\\
\texttt{0} & \texttt{0} & \texttt{0} & \texttt{0} & \texttt{0} & \texttt{0} & \texttt{0} & \texttt{1} & \texttt{1} & \texttt{0} & \texttt{0} & \texttt{0} & \texttt{1} & \texttt{1} \\
\end{tabular}
\vspace{1.5ex}

\noindent Doccano is used as the annotation tool \cite{doccano}, and ELIT is used for the tokenization \cite{DBLP:journals/corr/abs-2109-03903}.
To measure the inter-annotator agreement, ten dialogues are double-annotated that show a high kappa score of 84.4\%.

\subsubsection{Pseudo Annotation}
\label{ssec:pseudo-annotation}

\noindent Because our annotated data are relatively small, a larger dataset is pseudo-created for this task using 2,400 dialogues in the Switchboard \cite{stolcke-etal-2000-dialogue} and 6,808 dialogues in the BlendedSkillTalk \cite{smith-etal-2020-put} datasets (thus, a total of 9,208 dialogues).
These two datasets are chosen because their dialogues sound more speech-originated than others, having an adequate amount of filler words.
Among the four types of diarization errors (Figure~\ref{fig:diarization-errors}), the ones caused by filler words (33\%) can be simulated on dialogues that do not contain such errors using statistical heuristics.\footnote{Filler words are inferred by the outputs of the part-of-speech tagger and the dependency parser in ELIT.}

The errors associated with filler words are pseudo-inserted into dialogues from the two datasets by finding an utterance either beginning or ending with a filler word and concatenating it with an utterance before or next to it.
Global search is made to the entire dialogues for finding such utterances to mimic the distributions in Table~\ref{tab:filler-distribution} such that about 40.4\% of the dialogues in the pseudo-created data would contain two utterances with diarization errors, where 46.7\% of them are caused by the filler word \textit{okay}, and so on.
It is possible that more than two utterances get joined; in our case, up to 8 utterances are concatenated.
Table~\ref{tab:diarization-data} includes the statistics of our pseudo-created dataset for transfer learning.

\begin{table}[htbp!]
\caption{Distributions of filler words w.r.t. diarization errors. Dist: percentage of dialogues containing \# number of utterances with errors caused by the filler words. \textit{filler\_word}: percentage of the filler word appearing in the corresponding dialogue group.}
\centering\small{ 
\begin{tabular}{cc|ccccccccc} 
\toprule
\bf \# & \bf Dist & \it okay & \it yeah & \it right & \it um & \it so & \it uh & \it well & \it like & \it oh \\
\midrule
2 &      40.4 & 46.7 & 16.0 & 8.0 & 8.5 & $\:\:$8.0 & 4.4 & 4.7 & 0.2 & 3.1 \\
3 &      35.9 & 33.3 & 29.8 & 3.9 & 8.5 & 11.1 & 6.2 & 2.3 & 0.6 & 4.1 \\
4 & $\:\:$8.6 & 33.7 & 24.5 & 5.6 & 9.7 & 11.2 & 5.1 & 3.6 & 1.0 & 5.1 \\
5 & $\:\:$7.3 & 28.9 & 30.7 & 6.6 & 4.2 & 15.1 & 6.6 & 4.8 & 0.6 & 2.4 \\
\bottomrule
\end{tabular}}
\label{tab:filler-distribution}
\end{table}

\subsubsection{Joint Model}
\label{ssec:joint-diarization-model}

\noindent The joint model consists of two parts. First, we establish a binary classification task that enforces the model to learn to differentiate utterances that have diarization errors. 
The second part is a diarization model to tackle the problem specifically. 
The intention behind this design is that the binary classification task could enhance the embedding representation on a higher level to perform the diarization task better.

\begin{figure}[htbp!]
\centering
\includegraphics[width=0.55\columnwidth]{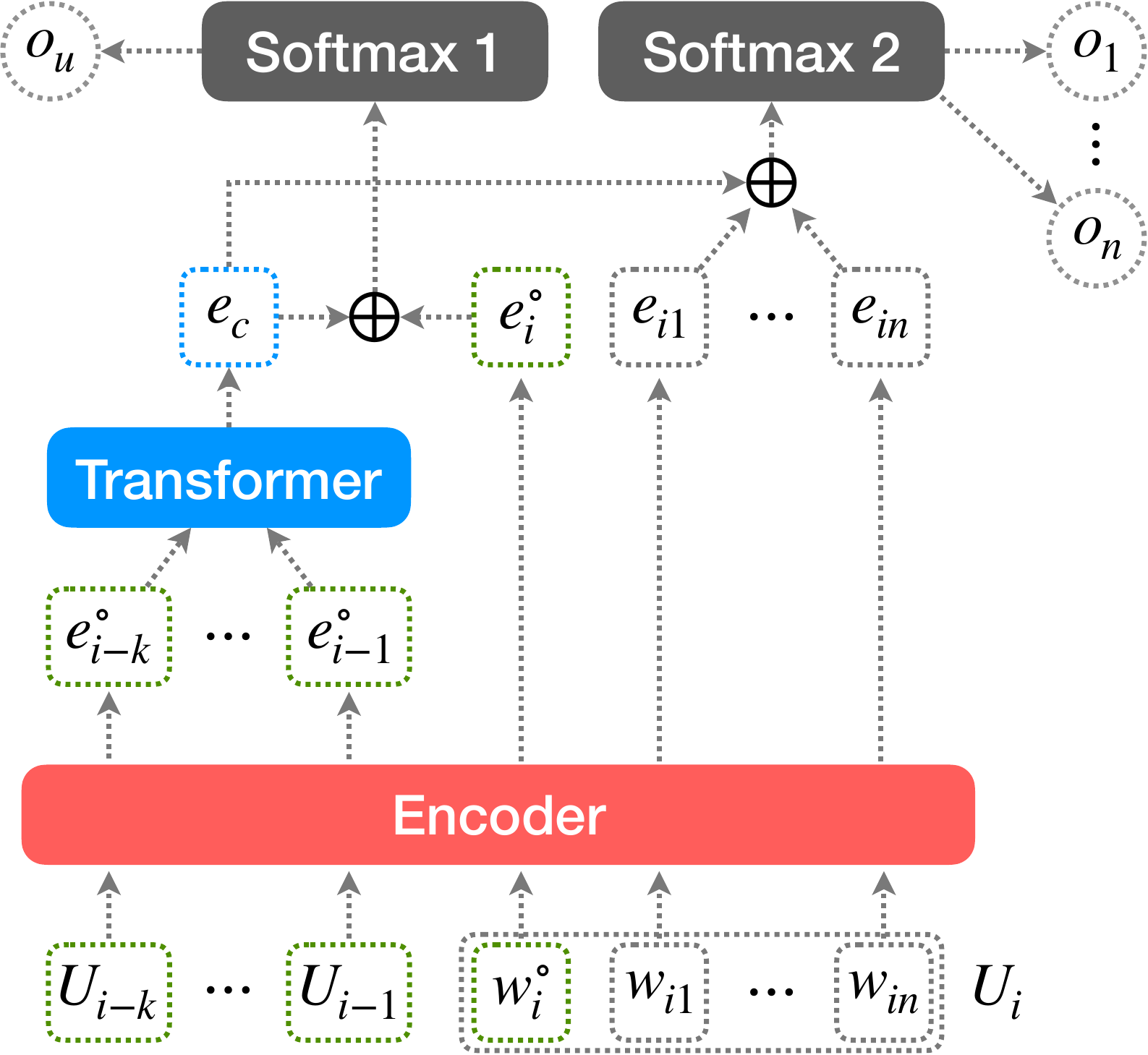}
\caption{The overview of our diarization model.}
\label{fig:diarization-model}
\end{figure}

Figure~\ref{fig:diarization-model} shows an overview of our speaker diarization model.
Let $U_i$ = $\{w_{i}^\circ, w_{i1}, .., w_{in}\}$ be the $i$'th utterance to be handled, where $w_{i}^\circ$ is the special token representing $U_i$ and $w_{ij}$ is the $j$'th token in~$U_i$.
$U_i$ is fed into the encoder $\mathcal{E}$ that generates the embeddings $\{e_{i}^\circ, e_{i1}, .., e_{in}\}$.
The previous utterances $\{U_{i-k}, .., U_{i-1}\}$ are also fed into $\mathcal{E}$ that generates $\{e_{i-k}^\circ, .., e_{i-1}^\circ\}$ (in our case, $k = 5$ that is the context window).
These embeddings are fed into a transformer layer for utterance-level weighting, which creates the context embedding $e_c$.
Finally, $e_c \oplus e_i^\circ$ is fed into a softmax layer that outputs $o_u$ to make a binary decision of whether or not $U_i$ includes any error.
Jointly, each $e_c \oplus e_{ij}$ is fed into another softmax that outputs $o_j$ to decide whether or not $w_{ij}$ is one of the last two tokens of an utterance.

\subsection{Dialogue Generation}
\label{sec:generation-model}

\begin{figure*}[htbp!]
\centering
\includegraphics[width=\textwidth]{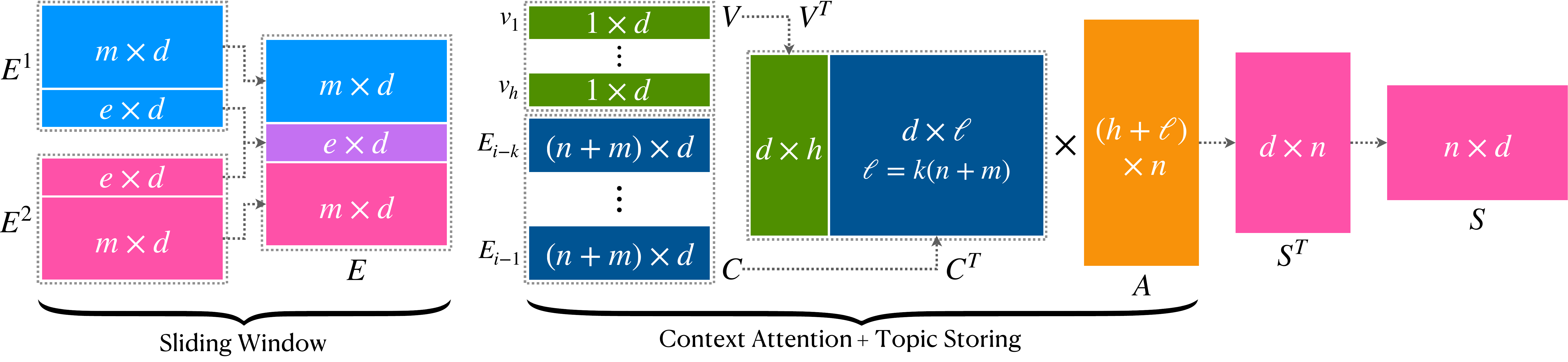}
\caption{The overview of our dialogue generation model.}
\label{fig:generation-model}
\end{figure*}

\noindent Figure~\ref{fig:generation-model} depicts an overview of our dialogue generation model.
Since inputs to the encoder $\mathcal{E}$ and the decoder $\mathcal{D}$ are limited by the total number of tokens that the pretrained language model accepts, \textit{sliding window} (Sec.~\ref{ssec:sliding-window}) and \textit{context attention} (Sec.~\ref{ssec:context-attention}) are proposed to handle long utterances and contexts in the previous utterances, respectively.
In addition, \textit{topic storing} is used to remember user-oriented topics brought up during the interview (Sec.~\ref{ssec:topic-storing}).
The input to $\mathcal{E}$ and output of $\mathcal{D}$ include the speaker ID \texttt{S1}, \texttt{S2} or special tokens to indicate the beginning \texttt{B}, the ending \texttt{E} and topic questions \texttt{Q} as the first token followed by an utterance from the interviewer or interviewee, respectively. 
Hyperparameters are finetuned by cross-validations.

\subsubsection{Sliding Window}
\label{ssec:sliding-window}

\noindent The sliding window technique aims to overcome the limitation of input length by separating a long sentence into multiple sections within between. The mathematical formulations are described below.
Let $n = m + e$ be the max-number of tokens that $\mathcal{E}$ and $\mathcal{D}$ accept ($e < m < n$).
Every utterance $U$ whose length is greater than $n$ is split into $U^1$ and $U^2$ as follows ($w_i$ is the $i$'th token in $U$):

\begin{equation*}
\begin{split}
U^1 & = \{w_1, \ldots, w_m, w_{m+1}, \ldots, w_{n}\} \\
U^2 & = \{w_{m+1}, \ldots, w_{n}, w_{n+1}, \ldots, w_{n+m}\} \\
\end{split}
\end{equation*}

\noindent In our case, $n$ = $128$, $m$ = $100$, and $e$ = $28$ such that $n+m = 228$ is sufficiently long enough to handle most utterances based on our stats.
$\mathcal{E}$ takes $U^1$ and $U^2$ then produces $E^1 = \{e^1_1, \ldots, e^1_{n}\}$ and $E^2 = \{e^2_{m+1}, \ldots, e^2_{n+m}\}$ where $e^*_i \in \mathbb{R}^{1 \times d}$ is the embedding of $w_i$.
Finally, the embedding matrix $E \in \mathbb{R}^{(n+m) \times d}$ of $U$ is created by stacking all of the following embeddings:

$$
\{e^1_1, \ldots, \frac{1}{2}\sum_{i=1}^2(e^{i}_{m+1}), \ldots, \frac{1}{2}\sum_{i=1}^2(e^{i}_{n}), \ldots, e^2_{n+m}\}
$$

\noindent For utterances whose lengths are less than or equal to $n$, zero-padding is used to transform $\mathcal{E}$'s output from $\mathbb{R}^{n \times d}$ to $\mathbb{R}^{(n+m) \times d}$.

\subsubsection{Context Attention}
\label{ssec:context-attention}

\noindent Let $U_i$ be the $i$'th utterance to be generated as output.
Let $C \in \mathbb{R}^{\ell \times d}$ be the context matrix stacking the embedding matrices of the previous utterances $\{E_{i-k}, .., E_{i-1}\}$, where $k$ is the number of previous utterances to be considered and $\ell = k(n+m)$. 
The transpose of $C$ is multiplied by the attention matrix $A \in \mathbb{R}^{\ell \times n}$ such that $C^T \cdot A \rightarrow S^T \in \mathbb{R}^{d \times n}$.
Thus, $S \in \mathbb{R}^{n \times d}$ represents the context summary of $U_{i-k}, .., U_{i-1}$, which is fed into the decoder $\mathcal{D}$.

\subsubsection{Topic Storing}
\label{ssec:topic-storing}

\noindent Even with the context attention, the model still has no memory of contexts prior to $U_{i-k}$, leading it to repeat the same topics that it has already initiated.
To overcome this issue, topic storage is introduced to remember key topics derived by the interviewer.
Every interview in our data came with 8-16 questions by the interviewer annotated after each interview by the data provider, who used those questions during the interview and thought they led to assessing crucial aspects of the interviewee.
Our final model considers these questions the “key topics” and dynamically stores them as the dialogue progresses.
During training, these questions are converted into embeddings and stored dynamically as a list of topics discussed in previous turns.
During decoding, the model generates such topical questions with a specific flag and stores them in the same way.

Let $Q = \{q_1, .., q_h\}$ be the topical question set.
During training, $\mathcal{D}$ learns to generate \texttt{Q} instead of \texttt{S1} as the first token of the interviewer's utterance that contains any $q_i \in Q$.
In addition, it generates \texttt{B}/\texttt{E} if the interviewer begins/ends the current dialogue with that utterance (Table~\ref{tab:interview-example}).
Any utterance starting with \texttt{Q} is encoded by $\mathcal{E}$ and Feed-forward layers that create abstract utterance embedding $v_i \in \mathbb{R}^{1 \times d}$ to represent topics.
These embeddings get stacked as the interview goes on to create the topic matrix $V \in \mathbb{R}^{h \times d}$.
If $|Q| < h$, then zero-padding is used to create $V$ (in our case, $h=16$).
Finally, $V$ is stacked with the context matrix $C$ (Sec.~\ref{ssec:context-attention}), and $(V \oplus C)^T \in \mathbb{R}^{d \times (h+\ell)}$ is multiplied by the attention matrix $A \in \mathbb{R}^{(h+\ell) \times n}$ to create the transpose of the context summary matrix $S \in \mathbb{R}^{n \times d}$.

\section{Results}
\label{sec:evaluation}

\subsection{Speaker Diarization Results}

\noindent Table~\ref{tab:diarization-data} shows the distributions of the pseudo-created data (Section~\ref{ssec:pseudo-annotation}), as well as our interview data (Section~\ref{sec:dataset}) before and after the diarization where errors in the train/dev/test sets are manually annotated (Section~\ref{ssec:manual-annotation}) and errors in the raw set are automatically corrected by the joint model (Section~\ref{ssec:joint-diarization-model}).
For the encoder, the RoBERTa large model is used \cite{DBLP:journals/corr/abs-1907-11692}.\footnote{Several transformer encoders including BERT \cite{devlin-etal-2019-bert} were evaluated and RoBERTa yielded the best results.}
After diarization, \texttt{S2}'s utterances with diarization errors get split such that the average length of \texttt{S2}'s utterances decreases while the average length of dialogues slightly increases.
Meanwhile, some parts of \texttt{S2}'s utterances, incorrectly separated from \texttt{S1}'s utterances by the transcriber, are recovered back to \texttt{S1}; thus, the average length of \texttt{S1}'s utterances increases.

\begin{table}[htbp!]
\caption{Distributions of the pseudo-created datasets (Switchboard, BST) and our interview data (before and after diarization). D: number of dialogues, U: avg-number of utterances, S1/S2: avg-number of tokens per utterance by \texttt{S1}/\texttt{S2}. \texttt{TRN}/\texttt{DEV}/\texttt{TST}: training/development/evaluation (annotated) sets. \texttt{RAW}: unannotated set. Note that we follow the same splits suggested by the original papers of the Switchboard and BST datasets for comparability.}
\centering\resizebox{\textwidth}{!}{
\begin{tabular}{c|rrrr|rrrr||rrrr|rrr}
\toprule
 & \multicolumn{4}{c|}{\bf Switchboard} & \multicolumn{4}{c||}{\bf BlendedSkillTalk (BST)} & \multicolumn{4}{c|}{\bf Interview Data (Before)} & \multicolumn{3}{c}{\bf (After)} \\
 & \multicolumn{1}{c}{\bf D} & \multicolumn{1}{c}{\bf U} & \multicolumn{1}{c}{\bf S1} & \multicolumn{1}{c|}{\bf S2} & \multicolumn{1}{c}{\bf D} & \multicolumn{1}{c}{\bf U} & \multicolumn{1}{c}{\bf S1} & \multicolumn{1}{c||}{\bf S2} & \multicolumn{1}{c}{\bf D} & \multicolumn{1}{c}{\bf U} & \multicolumn{1}{c}{\bf S1} & \multicolumn{1}{c|}{\bf S2} & \multicolumn{1}{c}{\bf U} & \multicolumn{1}{c}{\bf S1} & \multicolumn{1}{c}{\bf S2} \\
\midrule
\tt TRN & 1,115 & 42.2 & 31.6 & 31.3 & 4,819 & 6.2 & 25.0 & 25.2 &   140 & 42.2 & 37.4 & 73.1 & 43.8 & 39.3 & 64.0 \\
\tt DEV & 21 & 16.6 & 49.4 & 43.7 & 1,009 & 6.1 & 25.8 & 25.3 & 150 & 44.1 & 34.9 & 67.3 & 45.0 & 36.2 & 60.3 \\
\tt TST & 19 & 32.7 & 32.9 & 32.9 & 980 & 6.2 & 26.2 & 26.2 &   150 & 44.2 & 34.2 & 69.0 & 44.3 & 37.8 & 61.3 \\
\midrule
\tt RAW & \multicolumn{4}{c|}{N/A} & \multicolumn{4}{c||}{N/A} & 6,921 & 39.6 & 38.2 & 75.1 & 40.4 & 41.5 & 67.6 \\
\bottomrule
\end{tabular}}
\label{tab:diarization-data}
\end{table}

Table~\ref{tab:diarization-results} shows results of three models: the \textit{baseline} model taking $U_i$ and producing $O_w = \{o_1, .., o_n\}$, the \textit{context} model taking $\mathcal{U}_c = \{U_{i-k}, .., U_i\}$ and producing $O_u$, as well as the \textit{joint} model taking $\mathcal{U}_c$ and producing $O_u$ and $O_w$ (Figure~\ref{fig:diarization-model}).
The baseline model does not create $e_c$, so $e_{i*}$ are directly fed to Softmax 2. Also, the baseline and context models do not use $e_i^\circ$, so only Softmax 2 is used to produce the outputs.
For evaluation, the F1-scores of the label \texttt{1} on the last two tokens are used.
All models are developed three times and their average scores and standard deviations are reported.

\begin{table}[htbp!]
\caption{Diarization model performance. Ours: trained on \texttt{TRN} of our Interview data (After) in Table~\ref{tab:diarization-data}. Transferred: trained first on the \texttt{TRN} mixture of Switchboard and BST, then finetuned on \texttt{TRN} of our data.}
\centering{ 
\begin{tabular}{c|ccc}
\toprule
 & \multicolumn{1}{c}{\bf Baseline} & \multicolumn{1}{c}{\bf Context}  & \multicolumn{1}{c}{\bf Joint} \\
\midrule
Ours only   & 92.9$\pm$0.4 & 92.9$\pm$0.3 &     92.9$\pm$0.2 \\
Transferred & 93.2$\pm$0.3 & 93.4$\pm$0.3 & \bf 93.6$\pm$0.3 \\
\bottomrule
\end{tabular}}

\label{tab:diarization-results}
\end{table}

\noindent When trained on only our data, all models perform similarly.
The joint model slightly outperforms the others when transfer learning is applied.
Although the improvement is marginal, the joint model has a benefit of identifying utterances with diarization errors, showing the F1 score of 93.6\% for this task, while the transferred models generally show much higher performance on the other datasets than the non-transferred models.
Thus, the joint transferred model is used to auto-correct all dialogues in \texttt{RAW}.

\subsection{Dialogue Generation Results}

\noindent For our experiments, on the diarized data from the diarization model, the encoder and the decoder in BlenderBot 1.0 \cite{roller-etal-2021-recipes} are used.\footnote{There have been updated versions of BlenderBot introduced \cite{xu-etal-2022-beyond,https://doi.org/10.48550/arxiv.2208.03188}. However, we chose the first version for our experiments because we found it to be as effective yet much more efficient than the newer versions since the newer models focused on improvement on different perspectives, such as privacy and external knowledge incorporation.}
Three models are developed as follows:

\begin{itemize}
\item BB: Blenderbot Baseline Model
\item SW: Blenderbot with Sliding Window
\item CT: Blenderbot with Sliding Window and Concatenation of Topic Storing
\end{itemize}

\noindent All models are first trained on \texttt{raw} and finetuned on \texttt{TRN} in Table~\ref{tab:annotated-data}). We followed the setups on the training parameters in the original Blenderbot paper.
To assess real-life performance, ten interviews are conducted per model, where each interview consists of exactly 30 turns. 
Qualitative analysis is performed on the top-3 most frequently occurring errors as follows:

\begin{itemize}
\item Repetitions: how often it repeats topics already covered in the previous utterances.
\item Early Ending (EE): implies ending the interview without covering a sufficient amount of topics.
\item Off Topic (OT): how often it makes utterances that are not relevant to the current topic.
\end{itemize}

\noindent Table~\ref{tab:interviewbot-error-analysis} shows the error analysis results.
The repetition rates are significantly reduced as the model gets more advanced.
Compared to the baseline, the CT model conducts 3.5 times longer conversations before it attempts to end the interview while generating twice fewer off-topic utterances, which is very promising.
Examples of these error types are provided in Appendix~\ref{app:generation-errors}.

\begin{table}[htbp!]
\caption{The error analysis of all generation models. R: avg-\% of repeated topics, EE: avg-\% of the interview conducted before the model attempts to end (higher is better), OT: avg-\% of off-topic utterances.}
\centering\small{ 
\begin{tabular}{c|ccc}
\toprule
\bf Model & \multicolumn{1}{c}{\bf Repetitions (R)} & \multicolumn{1}{c}{\bf Early Ending (EE)} & \multicolumn{1}{c}{\bf Off Topic (OT)} \\
\midrule
BB & 30.0 & 13.3 & 20.0 \\
SW & 16.7 & 23.3 & 26.7 \\
CT & $\:\:$\bf 6.7 & \bf 46.7 & \bf 10.0 \\
\bottomrule
\end{tabular}
}

\label{tab:interviewbot-error-analysis}
\end{table}

\subsubsection{Static Evaluation}

\noindent Following previous work \cite{montahaei2019jointly}, static evaluation is performed on the CT model, where the input is every batch of $k$-utterances and prior topics per interview, and its output is compared to the corresponding human response in \texttt{TST} (Table~\ref{tab:annotated-data}).
The average \textsc{Bleu} score is 0.08 and cosine similarity is 0.19, which are low.
However, such static evaluation assesses each output independently and obstructs dialogue fluency by artificially inserting human utterances into the model, and thus, does not reveal its capability in conducting long contextualized interviews.


\subsubsection{Real-time Evaluation}

\noindent The CT model is deployed to an online text-based platform in a public cloud.
For real-time evaluation, five professional interviewers and ten students are invited to have conversations with our InterviewBot and give ratings from 1 to 5 to indicate their overall satisfaction. 
The average dialogue duration is 256 seconds. 
Almost half of the evaluators are satisfied (Scores 4 and 5) and another 40\% indicate a positive attitude on the coverage of topics and discussions (Score 3), implying that it performs reasonably well for this realistic setting (Table \ref{tab:interviewbot-rating-scores}).
Overall, with an average score of 3.5, the InterviewBot has shown great potential in applying to practical applications.

\begin{table}[htbp!]
\caption{The rating distribution of the InterviewBot conversations for real-time evaluation. 5: very satisfied, 4: satisfied, 3: neutral, 2: unsatisfied, 1: very unsatisfied.}

\centering\small{
\begin{tabular}{c|ccccc|c}
\toprule
    \bf Score & \bf 5 & \bf 4 & \bf 3 & \bf 2 & \bf 1 & \bf Average Score \\
\midrule
     \bf Interviewer (Count) & 1 & 1 & 2 & 1 & 0 & 3.4 \\
\midrule 
    \bf Student (Count) & 2 & 3 & 4 & 0 & 1 & 3.5 \\
\midrule \bf Total (Count) & 3 & 4 & 6 & 1 & 1 & 3.5 \\
\bottomrule
\end{tabular}
}
\label{tab:interviewbot-rating-scores}
\end{table}


\section{Discussion}
\label{sec:appendix}

\subsection{Chatbot Demonstration}

\noindent Table~\ref{tab:interview-example} presents an example dialogue conducted by our chatbot, showcasing the utilization of sliding window and topic storing (CT) techniques. 
Overall, the chatbot demonstrates its ability to conduct a comprehensive interview by asking relevant follow-up questions, adapting to various conversation topics, and providing meaningful responses accordingly.

\subsection{Examples of Diarization Errors}
\label{app:diarization-error-example}

\noindent The following are examples to illustrate the sources of diarization errors (\uwave{underlined}).
In many cases, interviewers and interviewees overlap in speeches or thinking out loud with or without filler words, which concatenates the two utterances.
A small portion of diarization errors are from speech recognition and word repetition errors.

\begin{itemize}
    \item \textbf{Arbitrary Concatenation} \\
    \textit{What do you think the benefits might be of this kind of technology? \uwave{If we develop it, I think this technology will eventually replace, um, human delivery.}}\vspace{0.5em}
    \item \textbf{Filler Words} \\
    \textit{Oh, no, I'm going to make majoring mathematics. \uwave{Okay. Okay.} Now why, why do you think receiving an education is important?}\vspace{0.5em}
    \item \textbf{Speech Recognition} \\
    \textit{Um, okay. My name is \uwave{<inaudible>}. I'm a senior year student come from \uwave{Hunger-Free}. Which high school are you from?}\vspace{0.5em}
    \item \textbf{Word Repetition}\\
    \textit{I heard it said, so it's kind of like a DIY \uwave{community community}. Are there community activities?}\vspace{0.5em}
\end{itemize}

\noindent We conduct speaker diarization errors by annotating 100 conversations with an average of 39 turns. The type of errors with their stats are shown in Table \ref{tab:speaker-diarization-error}. The major errors are caused by filler words and arbitrary concatenation. Counts of single errors are counted.
Table \ref{tab:error-type-example} listed examples underlined for each type of error, respectively.

\begin{table}[htbp!]
\caption{An interview dialogue conducted by our best model (CT in Section~\ref{sec:evaluation}). \texttt{S1/S2}: interviewer/interviewee (chatbot/human), \texttt{B/E}: beginning/ending utterance (chatbot), \texttt{Q}: topical question (chatbot).}
\centering\resizebox{\columnwidth}{!}{
\begin{tabular}{c|l} 
\toprule
\bf Flag & \multicolumn{1}{c}{\bf Utterance} \\
\midrule
\multirow[c]{2}{*}{\tt B}
 & Hello, and thank you so much for coming today. uh, Let us start by having you just to \\
 & state your full legal name and also the school that you are currently attending?\\
\midrule
\tt S2 & Sure, my name is David, and I go to Shanghai high school in China.\\
\midrule
\multirow[c]{2}{*}{\tt Q} & Great. um, yeah, {let's talk about leadership experiences.} Can you give me some specific \\
  & examples?\\
\midrule
\multirow[c]{2}{*}{\tt S2}
 & Sure, when I was a junior, I was elected to be the Vice President in the Student Union and \\
 & I organized many activities to help students to enrich their life on campus.\\
\midrule
\multirow[c]{2}{*}{\tt S1}
 & Okay, great. Thank you so much for sharing that. so, um, tell me a little bit about your \\
 & leadership style. What do you think you bring to the student union?\\
\midrule
\multirow[c]{3}{*}{\tt S2} & Sure, I am very open-minded, and with some of my colleagues, we together established \\
       & a lot of new programs on top of our existing ones. Many of those programs became very \\
       & popular. \\
\midrule
\multirow[c]{2}{*}{\tt S1} & Ok, great. and so, um, what, uh, what would you say has been the most challenging \\
 & aspect of being a leader? \\
\midrule
\multicolumn{2}{c}{$\cdots$} \\
\midrule
\multirow[c]{2}{*}{\tt Q}
 & Great. uh, um, hmm, let' talk about your courses. What courses did you take in high school \\
 & and which courses do you like most? \\
\midrule
\tt S2 & Sure, I took Math, English, PE, Chemistry, Art, and so on. I like Math most. \\
\midrule
\tt Q & Okay. so, um, tell me a little bit about your, your interest in mathematics. \\
\midrule
\multicolumn{2}{c}{$\cdots$} \\
\midrule
\multirow[c]{2}{*}{\tt E}
 & Okay, great. well, thank you very much for sharing that with me. um, and thank you for \\
 & taking the time out of your day to interview with us.\\
\bottomrule
\end{tabular}
}
\label{tab:interview-example}
\end{table}

\begin{table}[htbp!]
\caption{Speaker diarization error analysis in the text transcription from interview recordings. Notions are Misc: Miscellaneous; ASR: Auto Speech Recognition; WR: Word Repetition; FW: Filler Words; AC: Adjacent Concatenation.}
\centering{ 
\begin{tabular}{|c|c|c|c|}
\toprule
 {\bf Count} & {\bf Type} &  & \\
\midrule
212 & Misc & ASR: 147 & WR: 65 \\
289 & FW & & \\
378 & AC & & \\
\# of Convs & 100 & & \\
Avg \# of turns & 39 & & \\
\bottomrule
\end{tabular}}

\label{tab:speaker-diarization-error}
\end{table}

\begin{table}[htbp!]
\caption{Error type examples.}
\centering{}\resizebox{\columnwidth}{!}{
\begin{tabular}{c|l} 
\toprule
      \bf Error Type & \multicolumn{1}{c}{\bf Examples} \\
\midrule
\multirow{2}{*}{\tt ASR} & Um, okay. My name is \uwave{<inaudible>}. I'm a senior year student come from \\
 & \uwave{Hunger-Free} language school.\\ 
\midrule
\tt WR & I heard it said, so it's kind of like a DIY\uwave{community community}. \\
\midrule
\multirow{2}{*}{\tt FW} & Oh, no, I'm going to make majoring mathematics. \uwave{Okay. Okay.} Now why, \\
 & why do you think receiving an education is important?\\
\midrule
\multirow{2}{*}{\tt AC} & What do you think the benefits might be of this kind of technology? \uwave{If we develop} \\
 & \uwave{it, I think this technology will eventually replace, um, human delivery.}\\

\bottomrule
\end{tabular}
}
\label{tab:error-type-example}
\end{table}

\noindent From the table, there are different occasions for ASR errors, such as words not transcribed, words transcribed incorrectly, etc.
Word repetitions happen rarely when non-filler words repeat in a sentence. 
Concatenations of two adjacent utterances take the largest portion of all errors.

\subsection{Examples of Generation Limitations}
\label{app:generation-errors}

\noindent The following are examples for each type of generation error (\uwave{underlined}). 
An early ending prevents the Interviewbot from covering critical aspects of applicants and further discussion. 
The Interviewbot cannot name mentions properly, either skipping over or generating the wrong entities in the conversations.
Random Generation generates random words and punctuation and concatenates them incorrectly.
Not following up on previous topics hinders the Interviewbot from becoming proactively engaged with interviewees.

\begin{itemize}
    \item \textbf{Early Ending}\\
        InterviewBot: Could you please state your name?\\
        Interviewee: Sure. My name is David.\\
        ... 4 more turns ...\\
        InterviewBot: \uwave{Thank you and best wishes to your applications.}\vspace{0.5em}
    \item \textbf{Repetition}
        InterviewBot: Now why, why do you think receiving an education is important?\\
        ... more turns ...\\
        InterviewBot: Why do you think people value education so much?\vspace{0.5em}
    \item \textbf{Off Topic}
        Interviewee: Sure, I really like Mathemathcs.\\
        InterviewBot: \uwave{What do you do for fun?}
\end{itemize}


\section{Conclusion}
\label{sec:conclusion}

\noindent Our InterviewBot is a model-based dialogue system equipped with contextual awareness and topic sensitivity that conducts college admission interviews. 
Questions covering diverse topics and discussions in extended follow-ups are carried along the conversations, which have been assessed by professional interviewers and student volunteers. 
The average satisfaction score of 3.5 projects the prevailing deployment of the InterviewBot for thousands of college applicants, especially for international students. 

With promising future applications, however, the current version of the InterviewBot has two major limitations. 
First, the early ending in Table~\ref{tab:interviewbot-error-analysis} still happens, where an ending utterance gets generated after an insufficient amount of turns, in which case, the interview may not cover critical aspects of the applicant.
Second, the bot makes good follow-ups to various topics; however, it needs to derive deeper discussions with more details.

In future work, the main focus is to enrich the follow-up discussions on topics or sub-topics during interactions by training the InterviewBot with more structured topic-flow materials. This task would indirectly alleviate the early ending issue by deepening the discussions on certain topics.

\clearpage

\vspace{6pt} 



\authorcontributions{Conceptualization, Z.W. and J.C.; Methodology, Z.W. and J.C.; Software: Z.W.; Formal analysis, Z.W. and J.C.; Investigation, Z.W. and N.K.; Resources, Z.W. and N.K. and T.C.; Data Curation, Z.W.; Writing---Original Draft Preparation, Z.W. and J.C.; Supervision, J.C.; Project Administration, J.C.; Funding Acquisition, J.C. and T.C. All authors have read and agreed to the published version of the manuscript.}

\funding{This research was funded by InitialView.}

\institutionalreview{Not applicable}

\informedconsent{Not applicable}


\acknowledgments{We gratefully acknowledge the support of the InitialView grant.}

\conflictsofinterest{The authors declare no conflict of interest. The funding sponsors had no role in the design of the study; in the collection, analyses, or interpretation of data; in the writing of the manuscript, and in the decision to publish the results.}





\clearpage
%




\begin{adjustwidth}{-\extralength}{0cm}

\reftitle{References}


\bibliography{custom}

\end{adjustwidth}

\end{document}